\documentclass[10pt,twocolumn,letterpaper]{article}

\usepackage{cvpr}
\usepackage{times}
\usepackage{epsfig}
\usepackage{graphicx}
\usepackage{amsmath}
\usepackage{amssymb}

\usepackage[breaklinks=true,bookmarks=false]{hyperref}

\cvprfinalcopy

\begin{document}

\title{Deep Face Forgery Detection}

\author{
\parbox{4cm}{\centering Nika Dogonadze\quad}
\parbox{4cm}{\centering Jana Obernosterer\quad}
\parbox{4cm}{\centering Ji Hou}\\[0.3em]
Technical University of Munich \\
{\tt\small nika.dogonadze@tum.de} \quad {\tt\small jana.obernosterer@tum.de} \quad {\tt\small ji.hou@tum.de}
}

\maketitle

\begin{abstract}
   Rapid progress in deep learning is continuously making it easier and cheaper to generate video forgeries. Hence, it becomes very important to have a reliable way of detecting these forgeries. This paper describes such an approach for various tampering scenarios.
   The problem is modelled as a per-frame binary classification task. We propose to use transfer learning from face recognition task to improve tampering detection on many different facial manipulation scenarios. Furthermore, in low resolution settings, where single frame detection performs poorly, we try to make use of neighboring frames for middle frame classification. We evaluate both approaches on the public  \href{http://kaldir.vc.in.tum.de/faceforensics_benchmark/}{FaceForensics benchmark}, achieving state of the art accuracy. All the complementary code used for this paper is available on \href{https://github.com/Megatvini/DeepFaceForgeryDetection}{GitHub}\footnote{\href{https://github.com/Megatvini/DeepFaceForgeryDetection}{https://github.com/Megatvini/DeepFaceForgeryDetection}}.
\end{abstract}

\section{Introduction}

The overall objective of our project is to improve classification accuracy on manipulated portrait video frames. We limit our input data to portrait videos with a fairly constant background and focus on detecting manipulations resulting from one of four different face forgery methods. Given a pair of a source and a target video:

\begin{itemize}
    \item FaceSwap \cite{faceswap} - tries to smoothly transfer the face region, achieving facial identity manipulation. It is graphics-based and can be implemented efficiently on a CPU.
    \item DeepFakes \cite{deepfakes} - also manipulates facial identity, but is based on auto-encoder structures trained on reconstructing the source and the target faces. This is much less efficient, since each new face requires a separate neural network.
    \item Face2Face \cite{face2face} - maintains the identity of the target, but transfers facial expressions from the source. It uses key frame detection technique to build a dense representation of a face and then manipulate it to generate altered expressions.
    \item NeuralTextures \cite{NeuralTextures} - is a Generative Adversarial Network-based facial expression manipulation that only modifies the mouth region. In practice, this is proved to be the most difficult to detect, both for humans \cite{faceforensics++} and automated models.
\end{itemize}

All of these manipulation methods only alter the facial region of the visible person in the video. We model the detection objective as a per-frame binary classification task, classifying each frame of a given video as being either manipulated or pristine.

\subsection{Our Contribution}
We propose two main ideas for improving forgery detection. The first is using transfer learning from the face recognition domain. Since all the manipulations only alter the facial region of a person, this is an intuitive choice. 
Secondly, we try to take advantage of the sequential nature of a video,
considering temporal information to improve per-frame classification accuracy. Specifically, we want to use information from the previous $t$ and following $t$ frames for forgery detection on the current frame. We are motivated by the hypothesis that artifacts resulting from the manipulation might be easier to detect in some frames than in others and that by looking at multiple consecutive frames together and using the difference between them as an indicator we might be able to generalize our classification from obviously manipulated frames to more difficult examples. 

\subsection{Previous Work and Baseline}

Our work is mainly focused on improving the results of the FaceForensics++ paper~\cite{faceforensics++}. The authors published their code and most importantly the data~\cite{ff++data} on their GitHub~\cite{ff++github}. We were able to use their implementations as a starting point for our own approach to the problem. The paper also provides us with the final benchmark and baselines. 
It turns out that facial forgery detection poses a very difficult challenge when videos are highly compressed - especially, when the videos were manipulated using NeuralTextures. Thus, we decided to additionally focus our efforts on this particular case. 

\subsection{Data Set}

As the dataset used in the FaceForensics++ paper has been published online~\cite{ff++data}, we were able to use it for the training of our own model, thus ensuring a good comparability between the accuracies reported in the FaceForensics++ paper and our own results. The dataset consists of 1000 original videos from YouTube, each of which has been manipulated using the four manipulation methods described in the introduction resulting in a total of 4000 manipulated videos. Moreover, these 5000 videos are available in three different compression rates: raw (uncompressed), c23 (medium compression), c40 (high compression). For splitting the dataset into training, validation and test set, we use the official split published on the FaceForensics ++ GitHub~\cite{ff++github}, which splits the data up into a training set of 720 videos and 140 videos each for validation and testing.


\subsection{Preprocessing}
Data preprocessing included extracting frames from all videos in lossless .png-format and cropping each frame to only contain the facial region of the person visible. For the face detection we used the dlib library. All forgery methods applied to the dataset only manipulate the facial area. So, cropping away the negligible background drastically sped up our training and also improved performance.

\section{Single Frame Models}

\subsection{Baseline ResNet-18}
As a baseline, we took the very common convolutional architecture ResNet-18. We evaluate this baseline in two settings: First, starting from random weights and second, using weights pretrained on ImageNet and fine-tuning them on our data. 

\subsection{Inception ResNet V1}
We found that not using a pretrained model would always lead to overfitting and poor generalization. Even though pretraining on ImageNet was very helpful, the ImageNet classification task was also quite different from our task.  So, pretraining on a face recognition task would be a more natural choice. To test this assumption we tried an Inception ResNet V1 model pretrained on the VGGFace2 \cite{vggface2} face recognition dataset. This model also required to switch our face detection pipeline from dlib to MTCCN \cite{mtcnn}. 

\begin{figure}[t]
	\begin{center}
		\includegraphics[width=0.8\linewidth]{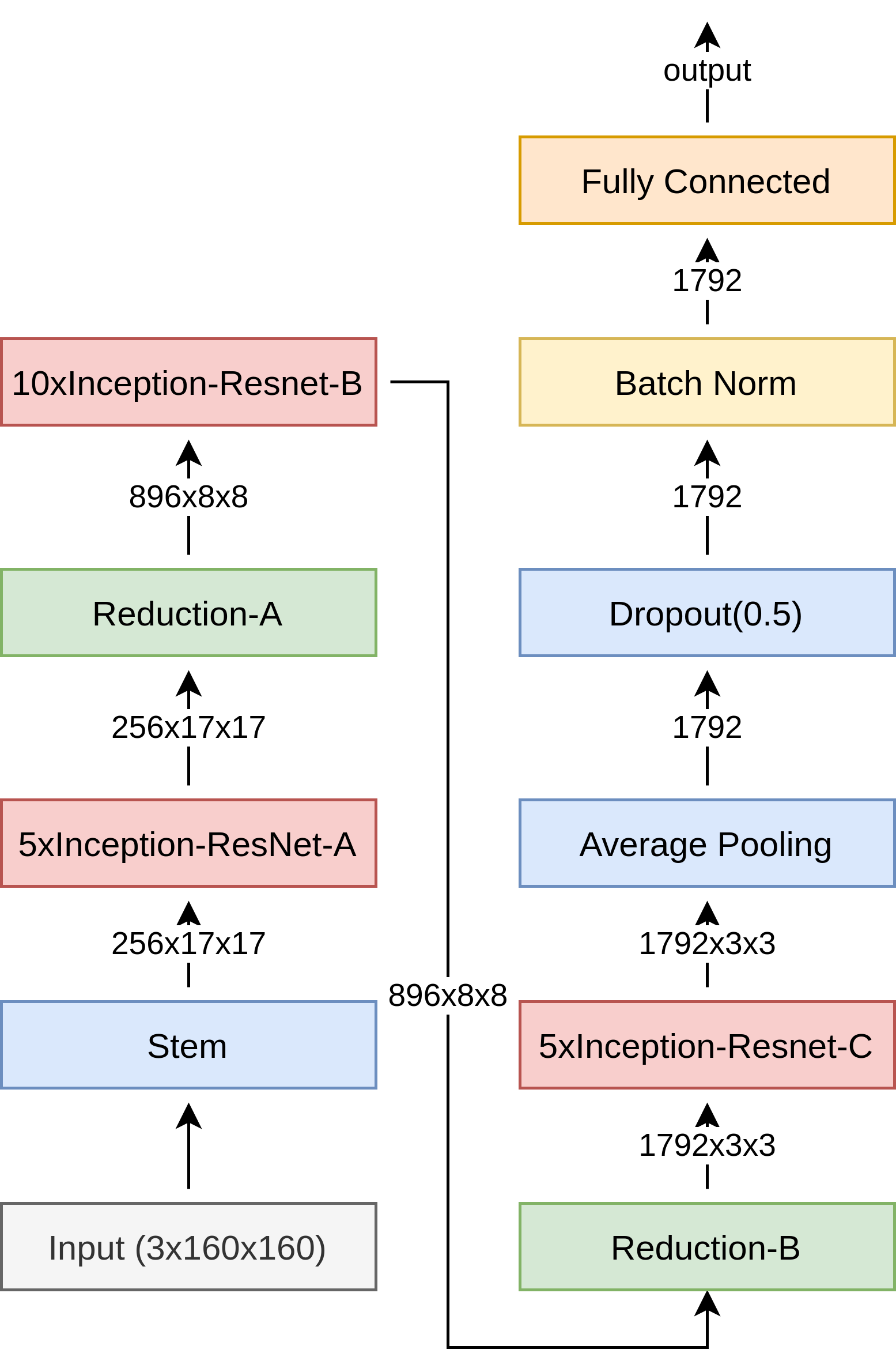}
	\end{center}
	\caption{Slightly modified architecture of the Inception ResNet V1 \cite{inception_resnet} model. This model was used for the final benchmark submission displayed in Figure \ref{fig:bechmark}.}
	\label{fig:inception_resnet}
\end{figure}

\section{Multi Frame Models}

\subsection{3D-CNN}

As a first attempt to include temporal information, we used a 3D convolutional model, namely a 3D ResNet-18 \cite{3dresnetpaper} for which a PyTorch implementation is available on GitHub \cite{3dresnetgithub}. Here we also saw the same overfitting problem when starting from non-pretrained weights. We tried to use a 3D ResNet pretrained on action recognition task to alleviate this problem. 

\begin{figure}[t]
	\begin{center}
		\includegraphics[width=1\linewidth]{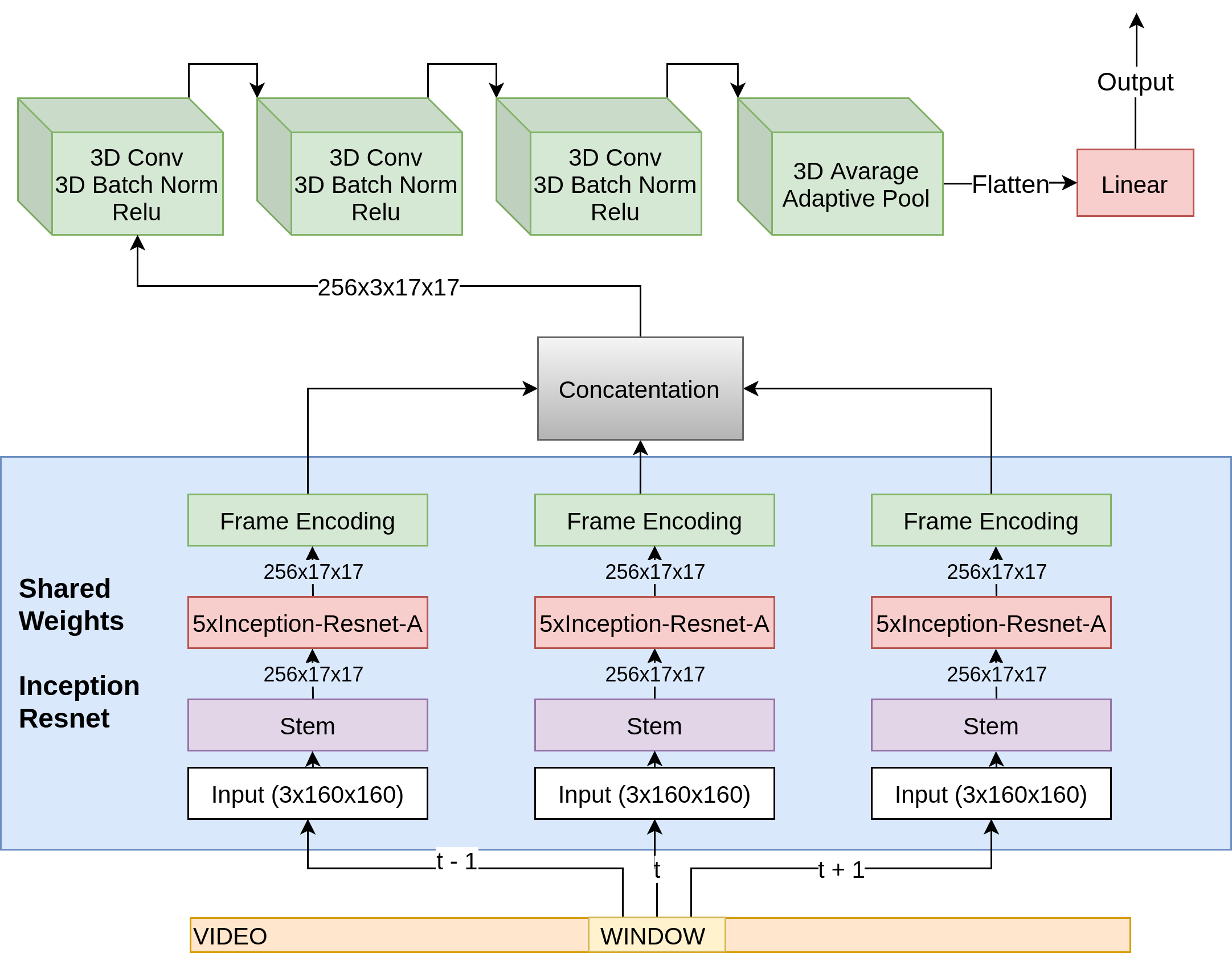}
	\end{center}
	\caption{Architecture of our 3D-CNN model. The encoder consists of a few initial layers from the Inception ResNet V1 model.
	The encoder uses the same shared weights for all frames. Number of total model parameters: 27,910,840.}
	\label{fig:cnn3d}
\end{figure}

The best accuracy was achieved by first using the first few layers of the 2D CNN encoder to get a feature map of each frame and then concatenate them to obtain a 3D input for the subsequent 3D convolutional model.

\subsection{Bi-LSTM}

\noindent

\begin{figure}[t]
	\begin{center}
		\includegraphics[width=0.9\linewidth]{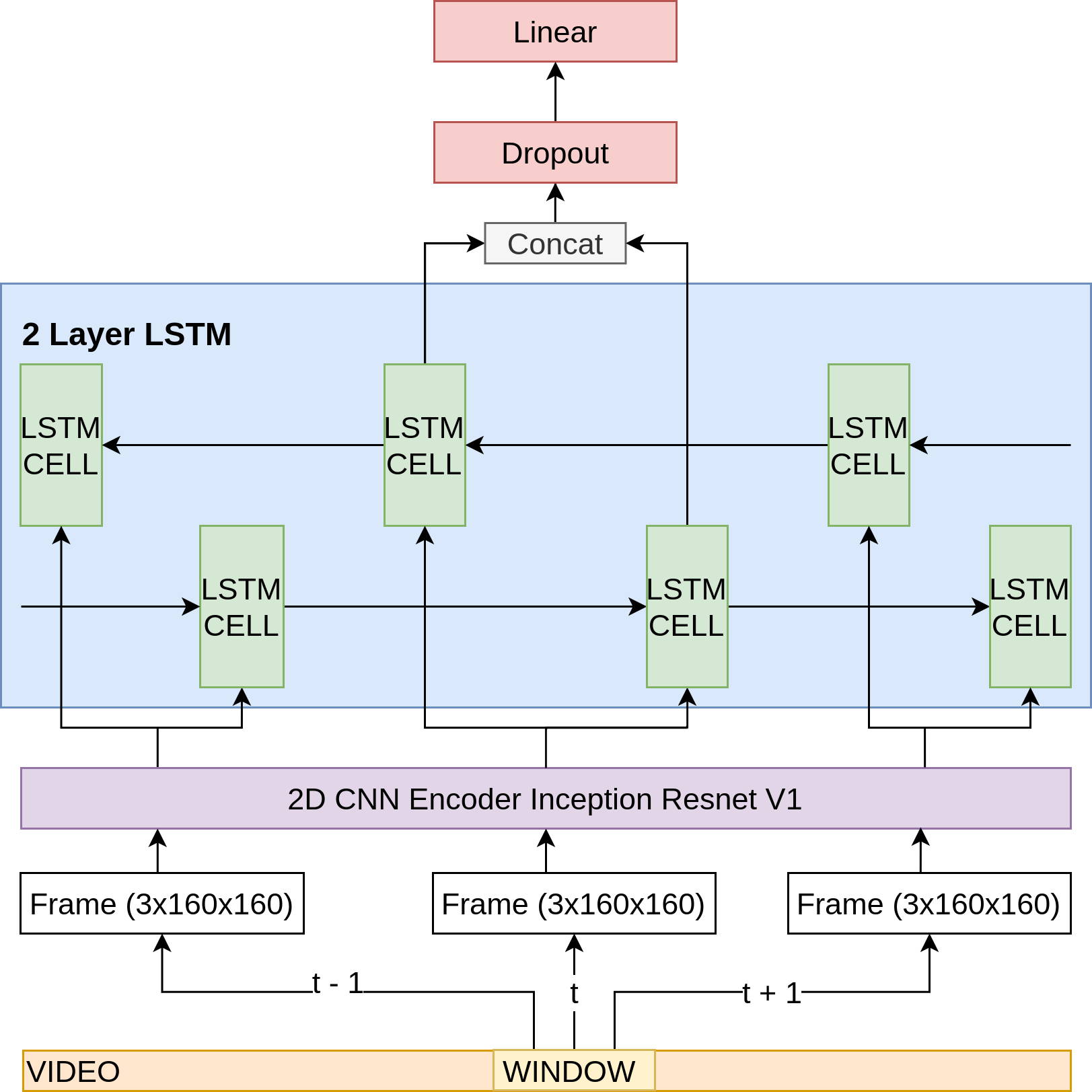}
	\end{center}
	\caption{Architecture of our Bi-LSTM model. The encoder uses the same shared weights for all frames. Number of total model parameters: 3,229,185.}
	\label{fig:bilstm}
\end{figure}

To capture the long-term dependencies between frames, we also used a recurrent model for classification, specifically, a Bi-directional Long Short Term Memory Network. We chose an LSTM in hope of being able to train on longer sequences without exploding computational costs and made it bi-directional to incorporate information about both previous and following frames when making a classification decision. A more detailed visualization of the architecture is shown in Figure \ref{fig:bilstm}. This architecture is inspired by ~\cite{lstmpaper}. Similar to our 3D convolutional network, the first layer has shared weights for all the frames in the window. It encodes each frame in the window into a 512-dimensional vector, generating a sequence of vectors for the Bi-LSTM to classify. 

\section{Evaluation of Results}
In this section we report the results of the various models discussed in the previous sections. All the results are evaluated using the official test split from the FaceForensics++ paper \cite{faceforensics++}. 

\subsection{Single Frame Classification}
In a single frame classification setting, our baseline is XceptionNet as published in the FaceForensics++ paper \cite{faceforensics++}. Table \ref{tab:single_frame} shows accuracies of our different models trained on the binary classification task between pristine images and NeuralTextures at the highest compression level (c40). Transfer learning improved the performance significantly, especially when using pretraining on the face recognition task.

\begin{table}
\begin{center}
\begin{tabular}{|l|c|c|c|}
\hline
Model & NT & Orig. & Avg \\
\hline \hline
ResNet-18 & 58.0 & 74.8 & 66.4 \\
ResNet-18 pretrained & 67.5 & 79.8 & 73.6 \\
InceptionResNet & 61.6 & 68.3 & 65.0\\
InceptionResNet pretrained & 75.3 & 74.2 & \textbf{74.8} \\
XceptionNet \cite{faceforensics++} & 80.7 & 52.4 & 66.5 \\

\hline
\end{tabular}
\end{center}
\caption{Accuracy results for various models trained and evaluated on classifying NeuralTextures vs. original frames at compression level c40. The ResNet model is pretrained on ImageNet classification. Inception ResNet is pretrained on VGGFace2 face recognition. }
\label{tab:single_frame}
\end{table}

\subsection{FaceForensics Benchmark}
\begin{figure*}
\begin{center}
\includegraphics[width=1\linewidth]{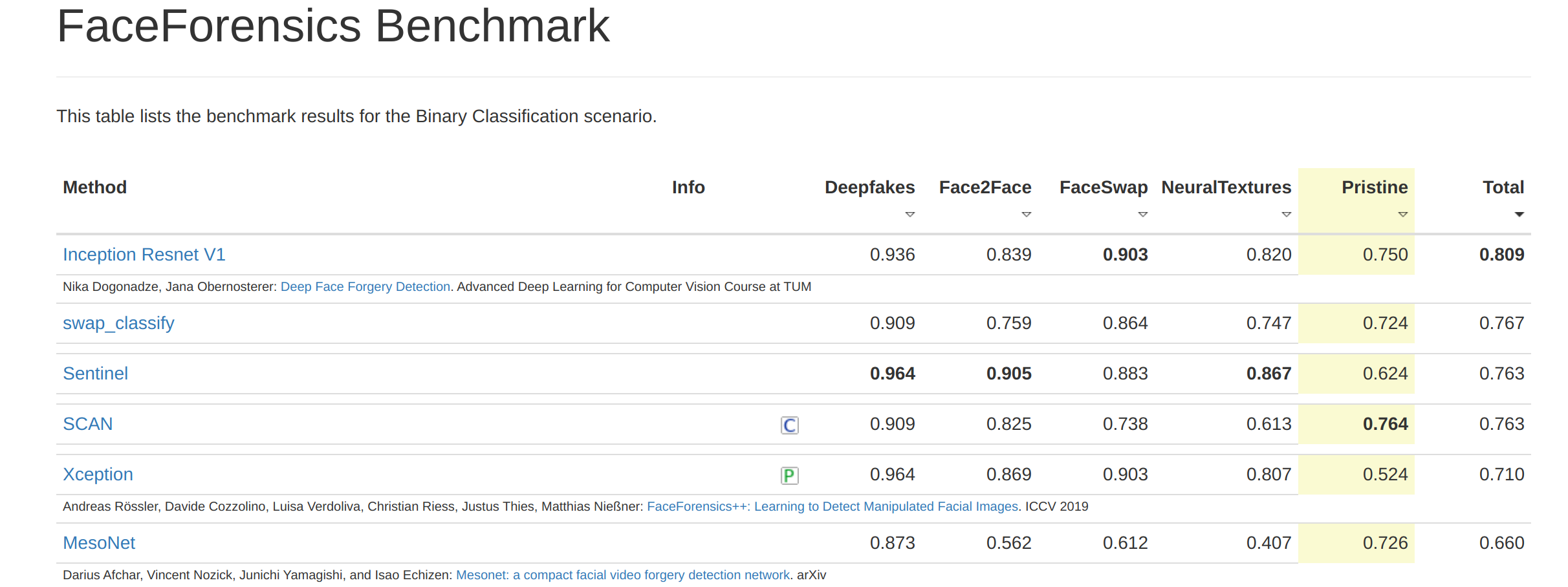}
\end{center}
   \caption{Our state of the art results on the FaceForensics Benchmark in direct comparison to the other leading submissions.  \textit{Screenshot taken on 01.04.2020}.}
\label{fig:bechmark}
\end{figure*}

\noindent
We also made a submission to the official \href{http://kaldir.vc.in.tum.de/faceforensics_benchmark/}{FaceForensics Benchmark} using our single frame model based on the Inception ResNet V1 model \cite{inception_resnet}. It was pretrained on the VGGFace2\cite{vggface2} face recognition task and further trained on a per frame binary classification on all four manipulation methods and all three compression levels. We achieved a new state of the art performance in just 12 hours of training on all 500GB of data, using Adam optimizer on a Nvidia Tesla V100 GPU. Details about the results are given in Table \ref{fig:bechmark}. They can also be seen on the official \href{http://kaldir.vc.in.tum.de/faceforensics_benchmark/}{benchmark website}.

\subsection{Multi-Frame Window Classification}
In a multi-frame setting we take a window of sequential frames from a video and classify only the middle frame. As all of the frames in a window are either pristine or tampered, only one classification decision has to be made. One obvious way to make this decision is to individually classify all the frames and then pick the majority vote. We call this a Majority Vote Model and use it as a baseline to make sure other multi-frame models learn something more than just classifying all the frames independently. See Table \ref{tab:seq_frame}. 

\subsection{3D CNN vs. Bi-LSTM}
Our first choice for incorporating multiple frames was to use 3D convolutional architecture as it seems reasonable to assume feature locality in temporal dimension for frames. The accuracy did improve over the baseline Majority Vote Model. However, we were able to achieve the same average $77.7$ accuracy using a Bi-LSTM model instead, with much fewer total parameters. Unfortunately, due to limited computational resources, we only trained and evaluated multi-frame models for window size of 7.

\subsection{Generalization from NeuralTextures}
Training only on highly compressed NeuralTextures vs. Pristine videos of a single-frame model did not generalize well to other forgery methods. However, it still worked well on moderately compressed (c23) or uncompressed (raw) videos of the same NeuralTexture manipulations. Exact results are shown in Table \ref{tab:nt_evaluation_results}.

\begin{table}
\begin{center}
\begin{tabular}{|l|c|c|c|}
\hline
Model & NT & Orig. & Avg \\
\hline \hline
Majority Vote baseline & 76.6 & 75.2 & 75.9 \\
ResNet-3D-18 pretrained & 71.8 & 79.1 & 75.5 \\
Our 3D Conv. network & 74.0 & 81.3 & \textbf{77.7} \\
LSTM with 2D Conv. encoder & 71.4 & 84.1 & \textbf{77.7} \\

\hline
\end{tabular}
\end{center}
\caption{Accuracy results for various models trained and evaluated on classifying NeuralTextures vs. original sequences of frames at compression level c40. For the given example a window size of 7 is used, i.e. all models get an input consisting of 3 previous and 3 subsequent frames when classifying the middle frame.}
\label{tab:seq_frame}
\end{table}

\begin{table}
\begin{center}
\begin{tabular}{|l|c|c|c|}
\hline
Data & raw & c23 & c40 \\
\hline \hline
Pristine & 61.6 & 79.6 & 74.2 \\
\hline
Neural Textures & 83.6 & 85.2 & 75.3 \\
\hline
DeepFakes & 56.8 & 56.5 & 41.8 \\
\hline
Face2Face & 50.8 & 54.0 & 46.8 \\
\hline
FaceSwap & 29.4 & 31.4 & 28.3 \\
\hline
\end{tabular}
\end{center}
\caption{Accuracy results of the Inception ResNet model trained only on Neural Textures and original images at high compression rate (c40) and evaluated on all forgery methods and compression levels. Accuracy on c40 is always worse than c23, but the same does not hold true for c23 and raw compression rates.}
\label{tab:nt_evaluation_results}
\end{table}

\section{Conclusion and Future Work}
In this paper, we showed that even though a very reliable manipulation detection with very highly compressed videos remains a challenge, using temporal information from frames does indeed give better results. In addition, transfer learning from the face recognition domain proved to be very useful for this task, achieving a new state of the art on the FaceForensics Benchmark. Transfer learning is especially important for being able to detect new manipulation methods that frequently emerge and for which there is not much training data available yet.

Because of limited computational resources we were not able to further examine how increasing the window size would help both 3D convolutions and recurrent models in the multi-frame window classification setting. 
It would also be interesting to evaluate how well learning generalizes from one manipulation method to another.

\section{Acknowledgements}
All work described in this paper was done as part of a project for the lecture "Advanced Deep Learning for Computer Vision" at the Technical University of Munich. We would especially like to thank Prof. Matthias Nie{\ss}ner and Prof. Laura Leal-Taixe who both held the lecture and provided us with the theoretical backgrounds needed to deliver the results described in this paper.

{\small
\bibliographystyle{ieee_fullname}
\bibliography{main}

\begin{thebibliography}{10}\itemsep=-1pt

\bibitem{deepfakes}
Deepfakes github.
\newblock \url{https://github.com/deepfakes/faceswap}, Accessed: 2020-01-30.

\bibitem{ff++data}
Faceforensics++ dataset.
\newblock
  \url{https://github.com/ondyari/FaceForensics/blob/master/dataset/README.md},
  Accessed: 2020-01-30.

\bibitem{faceswap}
Faceswap github.
\newblock \url{https://github.com/MarekKowalski/FaceSwap/}, Accessed:
  2020-01-30.

\bibitem{vggface2}
Q. Cao, L. Shen, W. Xie, O.~M. Parkhi, and A. Zisserman.
\newblock Vggface2: A dataset for recognising faces across pose and age.
\newblock In {\em International Conference on Automatic Face and Gesture
  Recognition}, 2018.

\bibitem{3dresnetgithub}
Kensho Hara.
\newblock 3d resnets for action recognition (cvpr 2018).
\newblock \url{https://github.com/kenshohara/3D-ResNets-PyTorch}, Accessed:
  2020-01-30.

\bibitem{3dresnetpaper}
Kensho Hara, Hirokatsu Kataoka, and Yutaka Satoh.
\newblock Learning spatio-temporal features with 3d residual networks for
  action recognition.
\newblock {\em CoRR}, abs/1708.07632, 2017.

\bibitem{ff++github}
Andreas R\"ossler.
\newblock Faceforensics++ github.
\newblock \url{https://github.com/ondyari/FaceForensics}, Accessed: 2020-01-30.

\bibitem{faceforensics++}
Andreas R\"ossler, Davide Cozzolino, Luisa Verdoliva, Christian Riess, Justus
  Thies, and Matthias Nießner.
\newblock Faceforensics++: Learning to detect manipulated facial images.
\newblock In {\em ICCV 2019}, 2019.

\bibitem{NeuralTextures}
Justus Thies, Michael Zollh{\"o}fer, and Matthias Nie{\ss}ner.
\newblock Deferred neural rendering: Image synthesis using neural textures.
\newblock {\em ACM Transactions on Graphics 2019 (TOG)}, 2019.

\bibitem{face2face}
Justus Thies, Michael Zollh\"{o}fer, Marc Stamminger, Christian Theobalt, and
  Matthias Nie{\ss}ner.
\newblock Face2face: Real-time face capture and reenactment of rgb videos.
\newblock {\em Commun. ACM}, 62(1):96--104, Dec. 2018.

\bibitem{lstmpaper}
Dalin Zhang, Lina Yao, Xiang Zhang, Sen Wang, Weitong Chen, and Robert Boots.
\newblock Eeg-based intention recognition from spatio-temporal representations
  via cascade and parallel convolutional recurrent neural networks.
\newblock 08 2017.

\bibitem{mtcnn}
K. Zhang, Z. Zhang, Z. Li, and Y. Qiao.
\newblock Joint face detection and alignment using multitask cascaded
  convolutional networks.
\newblock {\em IEEE Signal Processing Letters}, 23(10):1499--1503, Oct 2016.

\bibitem{inception_resnet}
X. {Zhang}, S. {Huang}, X. {Zhang}, W. {Wang}, Q. {Wang}, and D. {Yang}.
\newblock Residual inception: A new module combining modified residual with
  inception to improve network performance.
\newblock In {\em 2018 25th IEEE International Conference on Image Processing
  (ICIP)}, pages 3039--3043, Oct 2018.

\end{thebibliography}
}

\end{document}